# Smoothing Multivariate Performance Measures

**Xinhua Zhang**
Department of Computing Science
University of Alberta
Alberta, T6G 2E8, Canada
xinhua2@ualberta.ca

**Ankan Saha**
Department of Computer Science
University of Chicago
Chicago, IL 60637, USA
ankans@cs.uchicago.edu

**S.V. N. Vishwanathan**
Department of Statistics and
Department of Computer Science
Purdue University, IN 47907, USA
vishy@stat.purdue.edu

## Abstract

A Support Vector Method for multivariate performance measures was recently introduced by Joachims (2005). The underlying optimization problem is currently solved using cutting plane methods such as SVM-Perf and BMRM. One can show that these algorithms converge to an $\epsilon$ accurate solution in $O\left(\frac{1}{\lambda\epsilon}\right)$ iterations, where $\lambda$ is the trade-off parameter between the regularizer and the loss function. We present a smoothing strategy for multivariate performance scores, in particular precision/recall break-even point and ROCArea. When combined with Nesterov's accelerated gradient algorithm our smoothing strategy yields an optimization algorithm which converges to an $\epsilon$ accurate solution in $O^*\left(\min\left\{\frac{1}{\epsilon}, \frac{1}{\sqrt{\lambda\epsilon}}\right\}\right)$ iterations. Furthermore, the cost per iteration of our scheme is the same as that of SVM-Perf and BMRM. Empirical evaluation on a number of publicly available datasets shows that our method converges significantly faster than cutting plane methods without sacrificing generalization ability.

## 1 Background and Introduction

Different kinds of applications served by machine learning algorithms have varied and specific measures to judge the performance of the algorithms. In this paper we focus on efficient algorithms for directly optimizing multivariate performance measures such as precision/recall break-even point (PRBEP) and area under the Receiver Operating Characteristic curve (ROCArea). Given a training set with $n$ examples $\mathcal{X} := \{(\mathbf{x}_i, y_i)\}_{i=1}^n$ where $\mathbf{x}_i \in \mathbb{R}^p$ and $y_i \in \{+1, -1\}$, Joachims (2005) proposed an elegant formulation for this problem which minimizes the following regularized risk:

$$\min_{\mathbf{w}} \; J(\mathbf{w}) = \frac{\lambda}{2}\|\mathbf{w}\|^2 + R_{\text{emp}}(\mathbf{w}). \quad (1)$$

Here $\frac{1}{2}\|\mathbf{w}\|^2$ is the regularizer, $\lambda > 0$ is a trade-off parameter and the empirical risk $R_{\text{emp}}$ for contingency table based multivariate performance measures is

$$R_{\text{emp}}(\mathbf{w}) = \max_{\mathbf{z}\in\{-1,1\}^n}\left[\Delta(\mathbf{z},\mathbf{y}) + \frac{1}{n}\sum_{i=1}^n \langle \mathbf{w}, \mathbf{x}_i\rangle (z_i - y_i)\right]. \quad (2)$$

Here, $\Delta(\mathbf{z}, \mathbf{y})$ denotes the multivariate discrepancy between the correct labels $\mathbf{y} := (y_1, \ldots, y_n)^\top$ and a candidate labeling $\mathbf{z}$ (Joachims, 2005), and $\langle \cdot, \cdot \rangle$ denotes the Euclidean dot product. In order to compute the multivariate discrepancy for the PREBEP, which is the main focus of our work, we need the false positive and false negative rates, which are defined as

$$b = \sum_{i\in\mathcal{P}} \delta(z_i = -1), \text{ and } c = \sum_{j\in\mathcal{N}} \delta(z_j = 1).$$

Here $\delta(x) = 1$ if $x$ is true and 0 otherwise, while $\mathcal{P}$ and $\mathcal{N}$ denote the set of indices of positive ($y_i = +1$) and negative ($y_i = -1$) examples respectively. Furthermore, let $n_+ = |\mathcal{P}|$, $n_- = |\mathcal{N}|$. With this notation in place, $\Delta(\mathbf{z}, \mathbf{y})$ for PRBEP is defined as $b/n_+$ if $b = c$ and $-\infty$ otherwise (Joachims, 2005).

ROCArea, on the other hand, measures how many pairs of examples are mis-ordered. Denote $m = n_+ n_-$. (Joachims, 2005) proposed using the following empirical risk, $R_{\text{emp}}$, to directly optimize the ROCArea:

$$\frac{1}{m}\max_{\mathbf{z}\in\{-1,1\}^m}\left[\sum_{i\in\mathcal{P}, j\in\mathcal{N}} \tfrac{1}{2}(1 - z_{ij}) + z_{ij}\mathbf{w}^\top(\mathbf{x}_i - \mathbf{x}_j)\right]. \quad (3)$$

The empirical risks in (2) and (3) are non-smooth and this leads to difficulties in solving (1). However, cutting plane methods such as SVM-Perf (Joachims, 2006) and BMRM (Teo et al., 2010) can handle such problems. At each iteration these algorithms only require a sub-gradient of $R_{\text{emp}}$, which can be efficiently computed by a *separation* algorithm with $O(n \log n)$ effort for both (2) and (3) (Joachims, 2005). One can show that cutting plane methods can find an $\epsilon$ accurate solution of (1) after computing $O(\frac{1}{\lambda\epsilon})$ sub-gradients (Teo et al., 2010). These rates are optimal and cannot be improved (Zhang et al., 2011a).

One possible approach to break the $\Omega(\frac{1}{\lambda\epsilon})$ barrier is to approximate (1) by a smooth function, which in turn can be efficiently minimized by using either an accelerated gradient method or a quasi-Newton method (Nesterov, 2005, 2007). This technique for non-smooth optimization was pioneered by Nesterov (2005). We now describe some relevant details. Some mathematical preliminaries can be found in (Zhang et al., 2011b, Appendix B).

### 1.1 Nesterov's Formulation[1]

Let $A$ be a linear transform. Assume that given a prescribed precision $\epsilon$, we can find a smooth function $g_\mu^*(A^\top \mathbf{w})$ with a Lipschitz continuous gradient such that $\left|R_{\text{emp}}(\mathbf{w}) - g_\mu^*(A^\top \mathbf{w})\right| \leq \epsilon/2$ for all $\mathbf{w}$. Here $\mu$ is a parameter depending on $\epsilon$ and the condition number of the problem. It is easy to see that

$$J_\mu(\mathbf{w}) := \frac{\lambda}{2}\|\mathbf{w}\|^2 + g_\mu^*(A^\top \mathbf{w}) \tag{4}$$

satisfies $|J_\mu(\mathbf{w}) - J(\mathbf{w})| \leq \epsilon/2$ for all $\mathbf{w}$. In particular, if we find a $\mathbf{w}'$ such that $J_\mu(\mathbf{w}') \leq \min_{\mathbf{w}} J_\mu(\mathbf{w}) + \epsilon/2$, then it follows that $J(\mathbf{w}') \leq \min_{\mathbf{w}} J(\mathbf{w}) + \epsilon$. In other words, $\mathbf{w}'$ is an $\epsilon$ accurate solution for (1).

If we apply Nesterov's accelerated gradient method (Nesterov, 1983) to $J_\mu(\mathbf{w})$, as shown in Appendix A, one can find an $\epsilon$ accurate solution to $J(\mathbf{w})$ by querying the gradient of $g_\mu^\star(A^\top \mathbf{w})$ for

$$O^*\left(\sqrt{D}\,\|A\| \min\left\{\frac{1}{\epsilon}, \frac{1}{\sqrt{\lambda\epsilon}}\right\}\right) \tag{5}$$

number of times (Nesterov, 2005). Here $\|A\|$ is the matrix norm of $A$, and $D$ is a geometric constant that depends solely on $g_\mu^*$ and is independent of $\epsilon$ or $\lambda$.

Compared with the $O(\frac{1}{\lambda\epsilon})$ rates of cutting plane methods, the $\frac{1}{\sqrt{\lambda\epsilon}}$ part in (5) is already superior. Furthermore, many applications require $\lambda \ll \epsilon$ and in this case the $\frac{1}{\epsilon}$ part of the rate is even better. Note cutting plane methods rely on $\frac{\lambda}{2}\|\mathbf{w}\|^2$ to stabilize each update, and so they often converge slowly when $\lambda$ is small (Do et al., 2009).

Although the above scheme is conceptually simple, the smoothing of the objective function in (1) has to be performed very carefully in order to avoid dependence on $n$, the size of the training set. The main difficulties are two-fold. First, one needs to obtain a smooth approximation $g_\mu^*(A^\top \mathbf{w})$ to $R_{\text{emp}}(\mathbf{w})$ such that $\sqrt{D}\,\|A\|$ is small (ideally a constant). Second, we need to show that computing the gradient of $g_\mu^*(A^\top \mathbf{w})$ is no harder than computing a sub-gradient of $R_{\text{emp}}(\mathbf{w})$. In the sequel we will demonstrate how both the above difficulties can be overcome. Before describing our scheme in detail we would like to place our work in context by discussing some relevant related work.

### 1.2 Related Work

Training large models by using variants of stochastic gradient descent has recently become increasingly popular (Bottou, 2008; Shalev-Shwartz et al., 2007). However, stochastic gradient descent can only be applied when the empirical risk is *additively decomposable*, that is, it can be written as the average loss over individual data points. Since the non-linear multivariate scores such as the ones that we consider in this paper are not additively decomposable, this rules out the application of online algorithms to these problems.

Traditionally, batch optimizers such as the popular Sequential Minimal Optimization (SMO) worked in the dual (Platt, 1998). Recently, there has been significant research interest in optimizers which directly optimize (1) because there are some distinct advantages (Teo et al., 2010). Chapelle (2007) observed that to find a $\mathbf{w}$ which generalizes well, one only needs to solve the primal problem to very low accuracy (*e.g.*, $\epsilon \approx 0.01$). In fact, Chapelle (2007) introduced the idea of smoothing the objective function to the machine learning community. Specifically, he proposed to approximate the binary hinge loss by a smooth Huber's loss and used the Newton's method to solve this smoothed problem. This approach yielded the best overall performance in the Wild Competition Track of Sonnenburg et al. (2008) for training binary linear SVMs on large datasets. A similar smoothing approach is proposed by Zhou et al. (2010), but it is also only for binary hinge loss.

However, the smoothing proposed by Chapelle (2007) for the binary hinge loss is rather ad-hoc, and does not easily generalize to (2) and (3). Moreover, a function can be smoothed in many different ways and (Chapelle, 2007) did not explicitly relate the influence of smoothing on the rates of convergence of the solver. In contrast, we propose principled approaches to overcome these problems.

Of course, other smoothing techniques have also been explored in the literature. A popular approach is to replace the nonsmooth max term by a smooth log-sum-exp approximation (Boyd & Vandenberghe, 2004). In the case of binary classification this approximation is closely related to logistic regression (Bartlett et al.,

[1] For completeness we reproduce technical details from Nesterov (2005) in Appendix A.

2006; Zhang, 2004), and is equivalent to using an entropy regularizer in the dual. However, as we discuss in Section 2.1.2 this technique has some undesirable properties.

### 1.3 Notation and Paper Outline

We assume a standard setup as in Nesterov (2005), and make a running assumption that all $\mathbf{x}_i$ reside in a Euclidean ball of radius $R$. In Section 2 we will discuss how the smoothing function $g^*_\mu(A^\top \mathbf{w})$ can be designed for (2) and (3). We will focus on efficiently computing the gradient of the smooth objective function in Section 3. Empirical evaluation is presented in Section 4, and the paper concludes with a discussion in Section 5.

## 2 Reformulating the Empirical Risk

In order to approximate $R_{\text{emp}}$ by $g^*_\mu$ we will write $R_{\text{emp}}(\mathbf{w})$ as $g^*(A^\top \mathbf{w})$ for an appropriate linear transform $A$ and convex function $g^*$ with domain $Q$. Let $d$ be a strongly convex function with modulus 1 defined on $Q$. Furthermore, assume $\min_{\boldsymbol{\beta} \in Q} d(\boldsymbol{\beta}) = 0$ and denote $D = \max_{\boldsymbol{\beta} \in Q} d(\boldsymbol{\beta})$. $d$ is called a *prox-function.* Set

$$g^\star_\mu = (g + \mu d)^\star.$$

Then, one can show that $g^\star_\mu(A^\top \mathbf{w})$ is smooth and its gradient is Lipschitz continuous with constant at most $\frac{1}{\mu} \|A\|^2$. Clearly,

$$|g^\star_\mu(A^\top \mathbf{w}) - R_{\text{emp}}(\mathbf{w})| < \mu D, \tag{6}$$

and by choosing $\mu = \epsilon / D$, we can guarantee the approximation is uniformly upper bounded by $\epsilon$.

There are indeed many different ways of writing $R_{\text{emp}}(\mathbf{w})$ as $g^*(A^\top \mathbf{w})$, but the next two sections will demonstrate the advantage of our design.

### 2.1 Contingency Table Based Loss

Letting $\mathcal{S}^k$ denote the $k$ dimensional probability simplex, we can rewrite (2) as:

$$\begin{aligned} R_{\text{emp}}(\mathbf{w}) &= \max_{\mathbf{z} \in \{-1,1\}^n} \Big[ \Delta(\mathbf{z}, \mathbf{y}) + \frac{1}{n} \sum_{i=1}^n \langle \mathbf{w}, \mathbf{x}_i \rangle (z_i - y_i) \Big] \\ &= \max_{\boldsymbol{\alpha} \in \mathcal{S}^{2^n}} \sum_{\mathbf{z} \in \{-1,1\}^n} \alpha_{\mathbf{z}} \left( \Delta(\mathbf{z}, \mathbf{y}) + \frac{1}{n} \sum_{i=1}^n \langle \mathbf{w}, \mathbf{x}_i \rangle (z_i - y_i) \right) \quad (7) \\ &= \max_{\boldsymbol{\alpha} \in \mathcal{S}^{2^n}} \frac{-2}{n} \sum_{i=1}^n y_i \langle \mathbf{w}, \mathbf{x}_i \rangle \left( \sum_{\mathbf{z}: z_i = -y_i} \alpha_{\mathbf{z}} \right) + \sum_{\mathbf{z} \in \{-1,1\}^n} \alpha_{\mathbf{z}} \Delta(\mathbf{z}, \mathbf{y}). \end{aligned}$$

Define $\beta_i = \sum_{\mathbf{z}: z_i = -y_i} \alpha_{\mathbf{z}}$, then it is not hard to show that $R_{\text{emp}}(\mathbf{w})$ can be further rewritten as

$$\max_{\boldsymbol{\beta} \in [0,1]^n} \left\{ \frac{-2}{n} \sum_{i=1}^n y_i \langle \mathbf{w}, \mathbf{x}_i \rangle \beta_i - g(\boldsymbol{\beta}) \right\} \quad \text{where} \tag{8}$$

$$g(\boldsymbol{\beta}) := - \max_{\boldsymbol{\alpha} \in \mathcal{A}} \sum_{\mathbf{z}} \alpha_{\mathbf{z}} \Delta(\mathbf{z}, \mathbf{y}). \tag{9}$$

Here $\mathcal{A}$ is a subset of $\mathcal{S}^{2^n}$ defined via $\mathcal{A} = \{\boldsymbol{\alpha} : \sum_{\mathbf{z}: z_i = -y_i} \alpha_{\mathbf{z}} = \beta_i \text{ for all } i\}$. Indeed, this rewriting only requires that the mapping from $\boldsymbol{\alpha} \in \mathcal{S}^{2^n}$ to $\boldsymbol{\beta} \in Q := [0,1]^n$ is surjective. This is clear because for any $\boldsymbol{\beta} \in [0,1]^n$, a pre-image $\boldsymbol{\alpha}$ can be constructed:

$$\alpha_{\mathbf{z}} = \prod_{i=1}^n \gamma_i, \quad \text{where} \quad \gamma_i = \begin{cases} \beta_i & \text{if } z_i = -y_i \\ 1 - \beta_i & \text{if } z_i = y_i. \end{cases}$$

Furthermore we can show $g(\boldsymbol{\beta})$ is convex on $\boldsymbol{\beta} \in [0,1]^n$, (see Appendix C of Zhang et al. (2011b) for a proof). Using (8) it immediately follows that $R_{\text{emp}}(\mathbf{w}) = g^\star(A^\top \mathbf{w})$ where $A$ is a $p$-by-$n$ matrix whose $i$-th column is $\frac{-2}{n} y_i \mathbf{x}_i$, and $g^*$ denotes the Fenchel dual of $g$.

#### 2.1.1 $\sqrt{D}\,\|A\|$ for our design

Let us choose the prox-function $d(\boldsymbol{\beta})$ as $\frac{1}{2} \|\boldsymbol{\beta}\|^2$. Then $D = \max_{\boldsymbol{\beta} \in [0,1]^n} d(\boldsymbol{\beta}) = \frac{n}{2}$. The norm of $A = \frac{-2}{n}(y_1 \mathbf{x}_1, \ldots, y_n \mathbf{x}_n)$ can be tightly upper bounded by $\frac{2}{n}\sqrt{n} R = \frac{2R}{\sqrt{n}}$. Hence

$$\sqrt{D}\,\|A\| \le \sqrt{\frac{n}{2}} \frac{2R}{\sqrt{n}} = \sqrt{2} R.$$

#### 2.1.2 Alternatives

It is illuminating to see how naive choices for smoothing $R_{\text{emp}}$ can lead to large values of $\sqrt{D}\,\|A\|$. For instance, by (7), $R_{\text{emp}}(\mathbf{w})$ can be written as $h^\star(B^\top \mathbf{w})$ where $h(\boldsymbol{\alpha}) = -n \sum_{\mathbf{z} \in \{-1,1\}^n} \Delta(\mathbf{z}, \mathbf{y}) \alpha_{\mathbf{z}}$ if $\boldsymbol{\alpha}_{\mathbf{z}} \in [0, n^{-1}]$ and $\sum_{\mathbf{z}} \alpha_{\mathbf{z}} = \frac{1}{n}$, and $\infty$ elsewhere. $B$ is a $p$-by-$2^n$ matrix whose $\mathbf{z}$-th column is $\sum_{i=1}^n \mathbf{x}_i (z_i - y_i)$. $h(\boldsymbol{\alpha})$ has exactly the same form as the matrix game objective in (Nesterov, 2005), and a natural choice of prox-function $d$ is the entropy $d(\boldsymbol{\alpha}) = \sum_{\mathbf{z}} \alpha_{\mathbf{z}} \ln \alpha_{\mathbf{z}} + \frac{1}{n} \log n + \log 2$. However one can show that in this case $\sqrt{D}\,\|A\|$ can be $\Omega(nR)$ which grows linearly with $n$, the number of training examples. Similarly, the smoothing scheme proposed by Zhang et al. (2011a) also suffers from a linearly growing $\sqrt{D}\,\|A\|$.

Conceptually the key difficulty arises because the entropy $d$ is defined on a $2^n$ dimensional simplex. However, one can bypass the $\Omega(nR)$ dependence when $\Delta$ is additively decomposable. For example, if $\Delta(\mathbf{z}, \mathbf{y}) = \frac{1}{n} \sum_i \delta(z_i \neq y_i)$ in (2), then one can define $d(\boldsymbol{\alpha}) = \sum_i \alpha_i \log \alpha_i + (1 - \alpha_i) \log(1 - \alpha_i)$. By a straightforward derivation (omitted for brevity), one can show that $g^\star_\mu(A^\top \mathbf{w})$ recovers the logistic loss with its slope controlled by $\mu$, and hence $\sqrt{D}\,\|A\|$ is constant. However, since our $\Delta$ is not decomposable, the log-sum-exp approximation to (2) is not advantageous.

### 2.2 ROCArea

We rewrite $R_{\text{emp}}(\mathbf{w})$ from (3) as:

$$\frac{1}{m}\max_{\boldsymbol{\alpha}\in\mathcal{S}^{2^m}}\sum_{\mathbf{z}\in\{-1,1\}^m}\alpha_{\mathbf{z}}\left[\sum_{i\in\mathcal{P}}\sum_{j\in\mathcal{N}}\frac{1}{2}(1-z_{ij})+z_{ij}\mathbf{w}^\top(\mathbf{x}_i-\mathbf{x}_j)\right]$$
$$=\frac{1}{2}+\frac{1}{m}\max_{\boldsymbol{\alpha}}\left[-\frac{1}{2}\sum_{i,j}\left(\sum_{\mathbf{z}}z_{ij}\alpha_{\mathbf{z}}\right)\right.\qquad(10)$$
$$\left.+\sum_{i,j}\mathbf{w}^\top(\mathbf{x}_i-\mathbf{x}_j)\left(\sum_{\mathbf{z}}z_{ij}\alpha_{\mathbf{z}}\right)\right].$$

Let us define $\beta_{ij}=\sum_{\mathbf{z}}z_{ij}\alpha_{\mathbf{z}}$ for all $(i,j)\in\mathcal{P}\times\mathcal{N}$. This yields a compact form of $R_{\text{emp}}(\mathbf{w})$:

$$\frac{1}{2}+\frac{1}{m}\max_{\boldsymbol{\beta}}\left[-\frac{1}{2}\sum_{i,j}\beta_{ij}+\sum_{i,j}\beta_{ij}\mathbf{w}^\top(\mathbf{x}_i-\mathbf{x}_j)\right].\quad(11)$$

Clearly $\beta_{ij}\in[-1,1]$. In fact, we can further show that the mapping from $\boldsymbol{\alpha}\in\mathcal{S}^{2^m}$ to $\boldsymbol{\beta}\in Q:=[-1,1]^m$ is surjective. For any $\boldsymbol{\beta}$, a (non-unique) pre-image $\boldsymbol{\alpha}$ is

$$\alpha_{\mathbf{z}}=\prod_{ij}\gamma_{ij},\quad\text{where}\quad\gamma_{ij}=\begin{cases}\frac{1}{2}(1+\beta_{ij}) & \text{if } z_{ij}=1\\ \frac{1}{2}(1-\beta_{ij}) & \text{if } z_{ij}=-1\end{cases}.$$

Ignoring $\frac{1}{2}$, and using (11) $R_{\text{emp}}(\mathbf{w})$ can be written as $g^\star(A^\top\mathbf{w})$ where

$$g(\boldsymbol{\beta})=\begin{cases}\frac{1}{2m}\sum_{i,j}\beta_{ij} & \text{if }\boldsymbol{\beta}\in[-1,1]^m\\ +\infty & \text{elsewhere}\end{cases},$$

and $A$ is a $p$-by-$m$ matrix whose $(ij)$-th column is $\frac{1}{m}(\mathbf{x}_i-\mathbf{x}_j)$ for all $(i,j)\in\mathcal{P}\times\mathcal{N}$.

#### 2.2.1 $\sqrt{D}\,\|A\|$ for our design

Choose prox-function $d(\boldsymbol{\beta})=\frac{1}{2}\sum_{i,j}\beta_{ij}^2$. By a simple calculation, $D=\max_{\boldsymbol{\beta}}d(\boldsymbol{\beta})=\frac{m}{2}$. We can upper bound the norm of $A$ by

$$\|A\|=\max_{\|\mathbf{w}\|=\|\boldsymbol{\beta}\|=1}\mathbf{w}^\top A\boldsymbol{\beta}$$
$$=\max_{\|\mathbf{w}\|=\|\boldsymbol{\beta}\|=1}\sum_{i,j}\frac{\beta_{ij}}{m}\mathbf{w}^\top(\mathbf{x}_i-\mathbf{x}_j)$$
$$\leq\max_{\|\boldsymbol{\beta}\|=1}\frac{2R}{m}\sum_{i,j}|\beta_{ij}|\leq 2Rm^{-\frac{1}{2}},$$

where the last step follows from the Cauchy-Schwarz inequality. Therefore

$$\sqrt{D}\,\|A\|\leq\sqrt{\frac{m}{2}}\cdot\frac{2R}{\sqrt{m}}=\sqrt{2}R.$$

## 3 Computing the Gradient Efficiently

The last building block required to make the whole scheme work is an efficient algorithm to compute the gradient of the smoothed empirical risk $g_\mu^\star(A^\top\mathbf{w})$. By the chain rule and Corollary X.1.4.4 of (Hiriart-Urruty & Lemaréchal, 1993), we have

$$\frac{\partial}{\partial\mathbf{w}}g_\mu^\star(A^\top\mathbf{w})=A\boldsymbol{\beta}^*,\quad\text{where}\qquad(12)$$
$$\boldsymbol{\beta}^*=\operatorname*{argmax}_{\boldsymbol{\beta}\in Q}\left\langle\boldsymbol{\beta},A^\top\mathbf{w}\right\rangle-g(\boldsymbol{\beta})-\mu d(\boldsymbol{\beta}).\qquad(13)$$

Two major difficulties arise in computing the above gradient: (13) can be hard to solve (*e.g.* in the case of contingency table based loss), and the matrix vector product in (12) can be costly (*e.g.* $O(n^2p)$ for ROCArea). Below we show how these operations can be performed in $O(n\log n)$ time.

### 3.1 Contingency table based loss

Since $A$ is a $p\times n$ dimensional matrix and $\boldsymbol{\beta}^*$ is a $n$ dimensional vector, the matrix vector product in (12) can be computed in $O(np)$ time. Below we focus on solving (13).

To take into account the constraints in the definition of $g(\boldsymbol{\beta})$, we introduce Lagrangian multipliers $\theta_i$ and the optimization in (13) becomes

$$g_\mu^\star(A^\top\mathbf{w})=\max_{\boldsymbol{\beta}\in[0,1]^n}\left\{\frac{-2}{n}\sum_{i=1}^n y_i\left\langle\mathbf{w},\mathbf{x}_i\right\rangle\beta_i-\frac{\mu}{2}\sum_{i=1}^n\beta_i^2\right.$$
$$\left.+\max_{\boldsymbol{\alpha}\in\mathcal{S}^{2^n}}\left[\sum_{\mathbf{z}}\alpha_{\mathbf{z}}\Delta(\mathbf{z},\mathbf{y})+\min_{\boldsymbol{\theta}\in\mathbb{R}^n}\sum_{i=1}^n\theta_i\left(\sum_{\mathbf{z}:z_i=-y_i}\alpha_{\mathbf{z}}-\beta_i\right)\right]\right\}$$
$$\Leftrightarrow\min_{\boldsymbol{\theta}\in\mathbb{R}^n}\left\{\max_{\boldsymbol{\alpha}\in\mathcal{S}^{2^n}}\sum_{\mathbf{z}}\alpha_{\mathbf{z}}\left[\Delta(\mathbf{z},\mathbf{y})+\sum_i\theta_i\delta(z_i=-y_i)\right]\right.$$
$$\left.+\max_{\boldsymbol{\beta}\in[0,1]^n}\sum_{i=1}^n\left(\frac{-\mu}{2}\beta_i^2-\left(\frac{2}{n}y_i\left\langle\mathbf{w},\mathbf{x}_i\right\rangle+\theta_i\right)\beta_i\right)\right\}$$
$$\Leftrightarrow\min_{\boldsymbol{\theta}\in\mathbb{R}^n}\left\{\max_{\mathbf{z}}\overbrace{\left[\Delta(\mathbf{z},\mathbf{y})+\sum_i\theta_i\delta(z_i=-y_i)\right]}^{:=q(\mathbf{z},\boldsymbol{\theta})}\right.\qquad(14)$$
$$\left.+\sum_{i=1}^n\underbrace{\max_{\beta_i\in[0,1]}\left[\frac{-\mu}{2}\beta_i^2-\left(\frac{2}{n}y_i\left\langle\mathbf{w},\mathbf{x}_i\right\rangle+\theta_i\right)\beta_i\right]}_{:=h_i(\theta_i)}\right\}.$$

The last step is because all $\beta_i$ are decoupled and can be optimized independently. Let

$$D(\boldsymbol{\theta}):=\max_{\mathbf{z}}q(\mathbf{z},\boldsymbol{\theta})+\sum_{i=1}^n h_i(\theta_i)\text{ and }\boldsymbol{\theta}^*=\operatorname*{argmin}_{\boldsymbol{\theta}}D(\boldsymbol{\theta}).$$

Given $\boldsymbol{\theta}^*$ and denoting $a_i=\frac{-2}{n}y_i\left\langle\mathbf{w},\mathbf{x}_i\right\rangle$, we can recover the optimal $\beta(\theta_i^*)$ from the definition of $h_i(\theta_i^*)$

as follows:

$$\beta_i^* = \beta_i(\theta_i^*) = \begin{cases} 0 & \text{if } \theta_i^* \geq a_i \\ 1 & \text{if } \theta_i^* \leq a_i - \mu \\ \frac{1}{\mu}(a_i - \theta_i^*) & \text{if } \theta_i^* \in [a_i - \mu, a_i] \end{cases} \quad (15)$$

So, the main challenge that remains is to compute $\boldsymbol{\theta}^*$. Towards this end, first note that[2]:

$$\nabla_{\theta_i} h_i(\theta_i) = -\beta_i(\theta_i) \text{ and}$$

$$\nabla_{\boldsymbol{\theta}} q(\mathbf{z}, \boldsymbol{\theta}) = \text{co}\left\{\delta_{\mathbf{z}} : \mathbf{z} \in \operatorname*{argmax}_{\mathbf{z}} q(\mathbf{z}, \boldsymbol{\theta})\right\}.$$

Here $\delta_{\mathbf{z}} := (\delta(z_1 = -y_1), \ldots, \delta(z_n = -y_n))^\top$ and co$(\cdot)$ denotes the convex hull of a set. By the first order optimality conditions $\mathbf{0} \in D(\boldsymbol{\theta}^*)$ which implies that

$$-\boldsymbol{\beta}^* \in \text{co}\left\{\delta_{\mathbf{z}} : \mathbf{z} \in \operatorname*{argmax}_{\mathbf{z}} q(\mathbf{z}, \boldsymbol{\theta})\right\}.$$

The next theorem characterizes $\boldsymbol{\theta}^*$.

**Property 1.** *There must exist a unique optimal solution $\boldsymbol{\theta}^*$ of (14). Furthermore, $\theta_i^* \in [a_i - \mu, a_i]$ and can be computed in $O(n \log n)$ time for PRBEP.*

The proof of the theorem is technical and relegated to Appendix D of Zhang et al. (2011b). The entire algorithm is described in detail in Appendix E therein.

### 3.2 ROCArea loss

For the ROCArea loss, given the optimal $\boldsymbol{\beta}^*$ in (13) one can compute

$$\frac{\partial}{\partial \mathbf{w}} g_\mu^\star(A^\top \mathbf{w}) = \frac{1}{m} \sum_{i,j} \beta_{ij}^*(\mathbf{x}_i - \mathbf{x}_j)$$

$$= \frac{1}{m}\left[\sum_{i \in \mathcal{P}} \mathbf{x}_i \underbrace{\left(\sum_{j \in \mathcal{N}} \beta_{ij}^*\right)}_{:=\gamma_i} - \sum_{j \in \mathcal{N}} \mathbf{x}_j \underbrace{\left(\sum_{i \in \mathcal{P}} \beta_{ij}^*\right)}_{:=\gamma_j}\right].$$

If we can efficiently compute all $\gamma_i$ and $\gamma_j$, then the gradient can be computed in $O(np)$ time.

Given $\beta_{ij}^*$, a brute-force approach to compute $\gamma_i$ and $\gamma_j$ takes $O(m)$ time. We exploit the structure of the problem to reduce this cost to $O(n \log n)$, thus matching the complexity of the separation algorithm in (Joachims, 2005). Towards this end, we specialize (13) to ROCArea and write

$$\max_{\boldsymbol{\beta}} \left(\frac{1}{m} \sum_{i,j} \beta_{ij} \mathbf{w}^\top(\mathbf{x}_i - \mathbf{x}_j) - \frac{1}{2m} \sum_{i,j} \beta_{ij} - \frac{\mu}{2} \sum_{i,j} \beta_{ij}^2\right).$$

Since all $\beta_{ij}$ are decoupled, their optimal value can be easily found:

$$\beta_{ij}^* = \text{median}\,(1, a_i - a_j, -1) \text{ where}$$

$$a_i = \frac{1}{\mu m}\left(\mathbf{w}^\top \mathbf{x}_i - \frac{1}{4}\right), \text{ and } a_j = \frac{1}{\mu m}\left(\mathbf{w}^\top \mathbf{x}_j + \frac{1}{4}\right).$$

Below we give a high level description of how $\gamma_i$ for $i \in \mathcal{P}$ can be computed; the scheme for computing $\gamma_j$ for $j \in \mathcal{N}$ is identical. We omit the details for brevity.

For a given $i$, suppose we can divide $\mathcal{N}$ into three sets $\mathcal{M}_i^+$, $\mathcal{M}_i$, and $\mathcal{M}_i^-$ such that

- $j \in \mathcal{M}_i^+ \implies 1 < a_i - a_j$, hence $\beta_{ij}^* = 1$
- $j \in \mathcal{M}_i \implies a_i - a_j \in [-1, 1]$, hence $\beta_{ij}^* = a_i - a_j$
- $j \in \mathcal{M}_i^- \implies a_i - a_j < -1$, hence $\beta_{ij}^* = -1$.

Then, clearly

$$\gamma_i = \sum_{j \in \mathcal{N}} \beta_{ij}^* = |\mathcal{M}_i^+| - |\mathcal{M}_i^-| + |\mathcal{M}_i|\, a_i - \sum_{j \in \mathcal{M}_i} a_j.$$

In order to identify the sets $\mathcal{M}_i^+$, $\mathcal{M}_i$, and $\mathcal{M}_i^-$, we first sort both $\{a_i : i \in \mathcal{P}\}$ and $\{a_j : j \in \mathcal{N}\}$. We then walk down the sorted lists to identify for each $i$ the first and last indices $j$ such that $a_i - a_j \in [-1, 1]$. This is very similar to the algorithm used to merge two sorted lists, and takes $O(n_- + n_+) = O(n)$ time and space. The rest of the operations for computing $\gamma_i$ can be performed in $O(1)$ time with some straightforward book-keeping. The overall complexity of our algorithm is dominated by the complexity of sorting the two lists, which is $O(n \log n)$.

## 4 Empirical Evaluation

We used 11 publicly available datasets and focused our study on two aspects: The reduction in objective value as a function of CPU time, and the generalization performance of the models obtained via the two schemes.

**Practical Considerations** Optimizing the smooth objective function $J_\mu(\mathbf{w})$ using the optimization scheme described in (Nesterov, 2005) requires estimating the Lipschitz constant of the gradient of the $g_\mu^*(A^\top \mathbf{w})$. Although it can be automatically tuned by, *e.g.* (Beck & Teboulle, 2009), extra costs are incurred which slows down the optimization empirically (Zhang et al., 2011b, Appendix G). Therefore, we chose to optimize our smooth objective function using L-BFGS, a widely used quasi-Newton solver (Nocedal & Wright, 2006). The L-BFGS code is obtained from `http://www.chokkan.org/software/liblbfgs/`, which is a C port of the original Fortran implementation of L-BFGS by Nocedal. The size of the L-BFGS buffer determines the number of past parameter and gradient displacement vectors that are used in the construction of the quasi-Newton direction. We set the buffer size to 6. Following Chapelle (2007) we set $\epsilon = 0.001$ and observed that the solution obtained with this approximation generalizes well in most cases.

[2] We abuse notation slightly and use $\nabla$ to denote both the gradient and sub-gradient

Table 1: Dataset statistics. $n$: #examples, $d$: #features, $s$: feature density.

| dataset | $n$ | $d$ | $s(\%)$ | dataset | $n$ | $d$ | $s(\%)$ | dataset | $n$ | $d$ | $s(\%)$ |
|---|---|---|---|---|---|---|---|---|---|---|---|
| adult9 | 32,561 | 123 | 11.28 | covertype | 522,911 | 6,274,932 | 22.22 | web8 | 45,546 | 579,586 | 4.24 |
| astro-ph | 62,369 | 99,757 | 0.077 | news20 | 15,960 | 7,264,867 | 0.033 | worm | 615,620 | 804 | 25 |
| aut-avn | 56,862 | 20,707 | 0.25 | real-sim | 57,763 | 2,969,737 | 0.25 | kdd99 | 4,898,431 | 127 | 12.86 |
| reuters-c11 | 23,149 | 1,757,801 | 0.16 | reuters-ccat | 23,149 | 1,757,801 | 0.16 | | | | |

We compared this scheme with BMRM[3], a state-of-the-art cutting plane method for optimizing multivariate performance scores which directly minimizes the non-smooth objective function $J(\mathbf{w})$ (Teo et al., 2010). We obtained the latest BMRM code from `http://users.rsise.anu.edu.au/~chteo/BMRM.html` and used default settings. For a fair comparison, our smoothed loss was implemented as a subroutine in BMRM and L-BFGS was added as an alternative solver to the BMRM framework. All our code was written in C++ and will be made publicly available.

**Datasets** Table 1 summarizes the datasets used in our experiments. `adult9`, `astro-ph`, `news20`, `real-sim`, `reuters-c11`, `reuters-ccat` are from the same source as in Hsieh et al. (2008). `aut-avn` classifies documents on auto and aviation and was obtained from `http://www.cs.umass.edu/~mccallum/data/sraa.tar.gz`. `covertype` is from UCI repository (Merz & Murphy, 1998). We divided the whole dataset into training, validation and test set in the same way as in (Teo et al., 2010). For all datasets we used the $\lambda$ which yielded the best generalization performance using their corresponding validation sets.

Due to lack of space, we will only present results for three representative datasets in this proceedings version of the paper. Complete results can be found in (Zhang et al., 2011b, Appendix F).

### 4.1 Results

**Optimizing the primal objective $J(\mathbf{w})$** In the first experiment we study the effect of $\mu$ on optimizing the primal objective $J(\mathbf{w})$. The choice of $\mu$ is dictated by two conflicting requirements. On the one hand the uniform deviation bound (6) suggests setting $\mu = \epsilon/D$. However, this estimate is very conservative because in (6) we use $D$ an upper bound on the prox-function. In practice, the quality of the approximation depends on the value of the prox-function around the optimum. On the other hand, as $\mu$ increases, the strong convexity of $g$ increases, and this makes $g^*_\mu$ and hence $J_\mu$ easier to optimize. We set $\hat{\mu} = \epsilon/D$ and let $\mu \in \{\hat{\mu}, 100\hat{\mu}, 1000\hat{\mu}\}$ and compare the performance of our scheme with BMRM in Figures 1 and 2.

It is clear that for all the values of $\mu$, optimizing the smoothed objective function converges significantly faster than BMRM. Furthermore, for $\mu = \hat{\mu}$ and $\mu = 100\hat{\mu}$ we obtained a solution which was at most $\epsilon$ distance away from the solution obtained by BMRM. Somewhat surprisingly, the optimization trajectories were near identical for $\mu = \hat{\mu}$ and $\mu = 100\hat{\mu}$ indicating that increasing the strong convexity of $g$ did not significantly impact the convergence rates. However, $\mu = 1000\hat{\mu}$ did converge significantly faster, but to a worse quality solution.

**Performance on Test Set** We also studied the evolution of the PRBEP and ROCArea performance on the test data. For this, we obtained the solution after each iteration, computed its performance on the test set, and plotted the results in Figures 3 and 4. Clearly, the intermediate models output by our scheme achieve comparable (or better) PRBEM scores and ROCArea in time orders of magnitude faster those generated by BMRM.

## 5 Conclusion and Discussion

The non-smoothness of the loss function is an important consideration for algorithms which employ the kernel trick (Schölkopf & Smola, 2002). This is because such algorithms typically operate in the dual, and the non-smooth losses lead to sparse dual solutions. In many applications such as natural language processing, the kernel trick is not needed because the input data is sufficiently high dimensional. However, now we are "stuck" with a non-smooth objective function in the primal. While a lot of past work has been devoted to solving this non-smooth problem, one must bear in mind that optimization is a means to an end in machine learning. In line with this philosophy, we proposed efficient smoothing techniques to approximate the non-smooth function. When combined with a smooth optimization algorithm, our technique outperforms state-of-the-art non-smooth optimization algorithms for multivariate performance scores not only in terms of CPU time but also in terms of generalization performance.

[3]For quadratic regularizers, BMRM and SVM-Perf are equivalent.

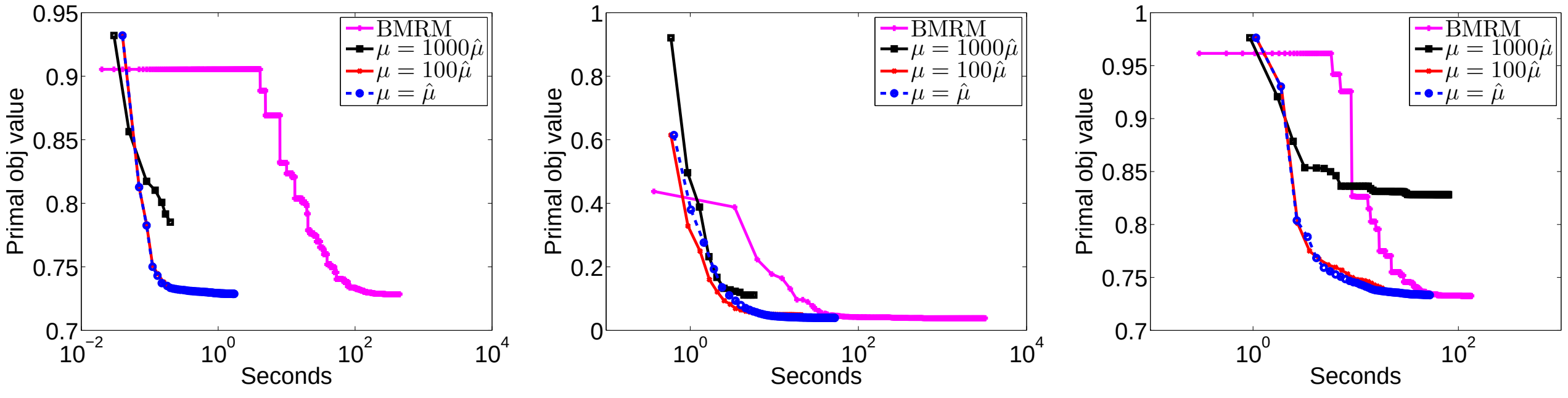

(a) adult9 (b) news20 (c) covertype

Figure 1: Primal objective versus CPU time for PRBEP.

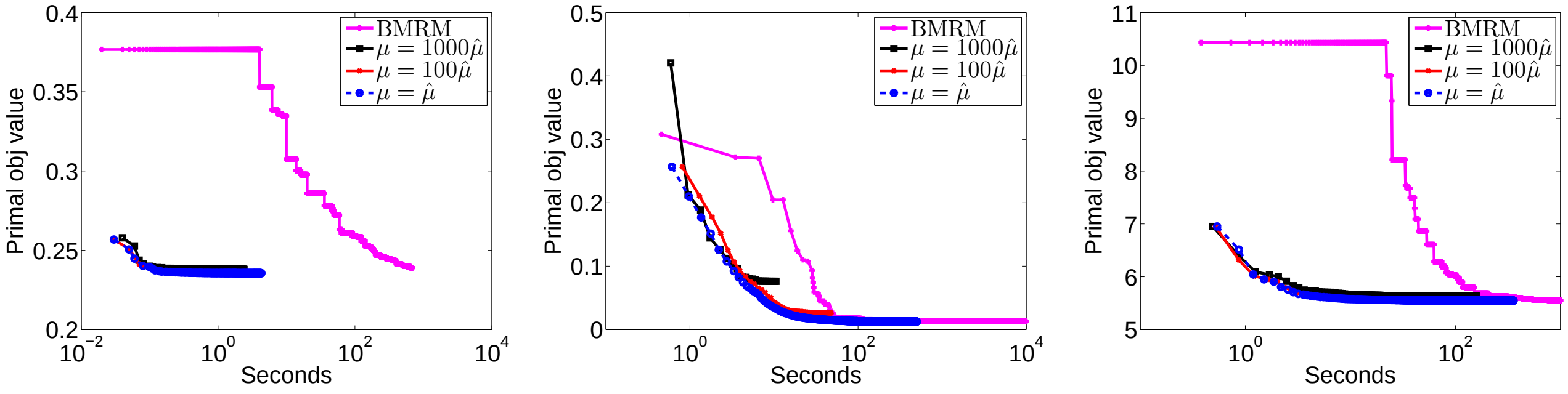

(a) adult9 (b) news20 (c) covertype

Figure 2: Primal objective versus CPU time for ROCArea.

It is also worthwhile noting that smoothing is not the right approach for every non-smooth problem. For example, although it is easy to smooth the $L_1$ norm regularizer, it is not recommended; the sparsity of the solution is an important statistical property of these algorithms and smoothing destroys this property.

In future work we would like to extend our techniques to handle more complicated contingency based multivariate performance measures such as the $F_1$-score. We would also like to extend smoothing to matching loss functions commonly used in ranking, where we believe our techniques will solve a smoothed version of the Hungarian marriage problem.

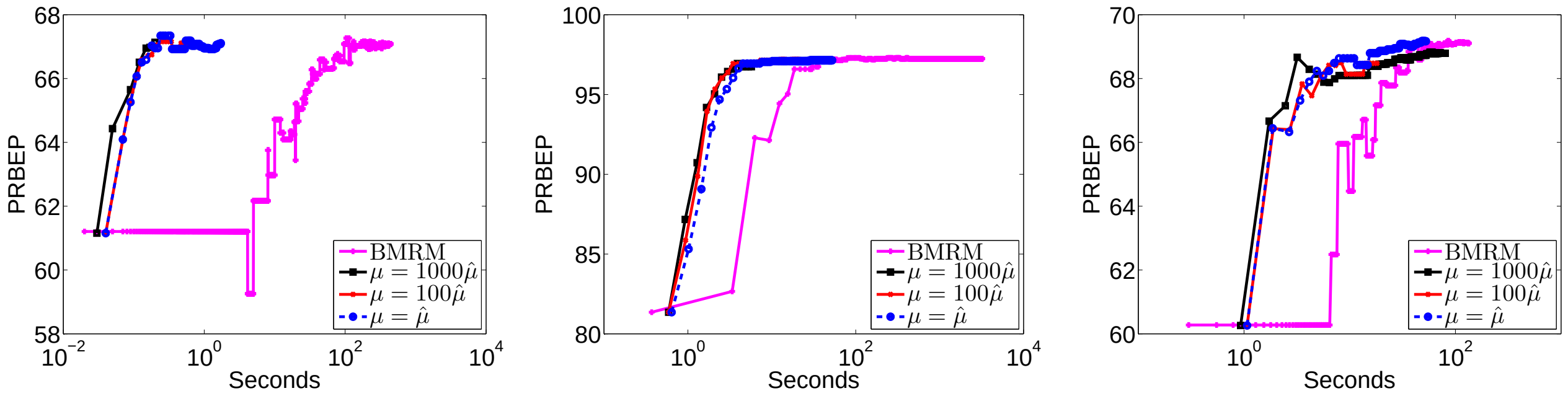

(a) adult9 (b) news20 (c) covertype

Figure 3: PRBEP on test data versus CPU time.

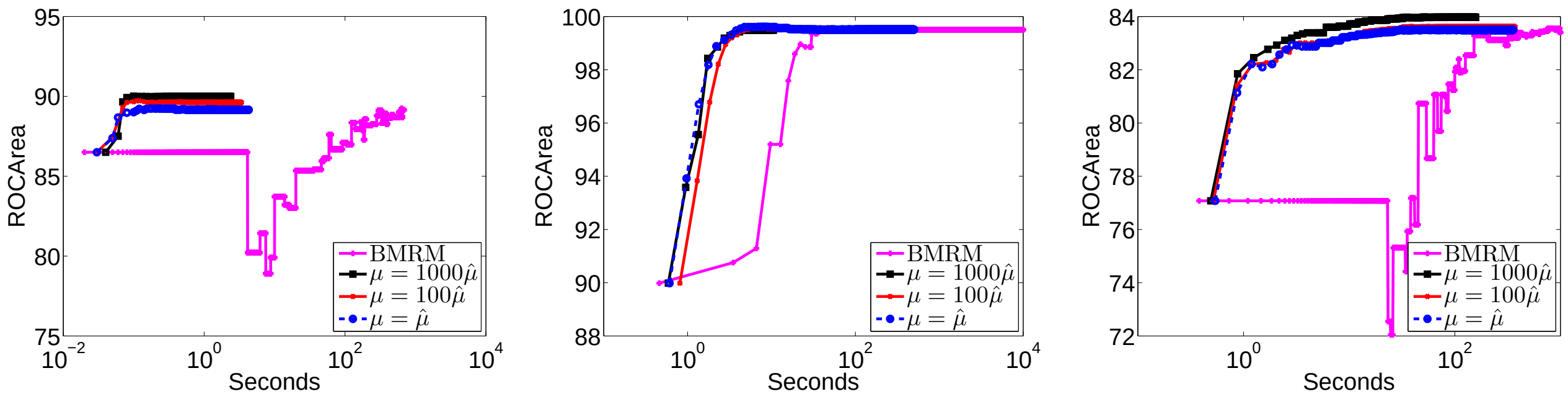

(a) adult9 (b) news20 (c) covertype

Figure 4: ROCArea on test data versus CPU time.

**Acknowledgements.** We thank the three anonymous reviewers for the very helpful comments and suggestions.

## A The Smoothing Procedure

The idea of the smoothing technique in (Nesterov, 2005) can be motivated by using the Theorem 4.2.1 and 4.2.2 in (Hiriart-Urruty & Lemaréchal, 1993).

**Lemma 1.** *If $f : \mathbb{R}^n \to \mathbb{R}$ is convex and differentiable, and $\nabla f$ is Lipschitz continuous with constant $L$ (called $L$-l.c.g), then $f^\star$ is strongly convex with modulus $\frac{1}{L}$ (called $\frac{1}{L}$-sc). Conversely, if $f : \mathbb{R}^n \to \mathbb{R} \cup \{\infty\}$ is $\sigma$-sc, then $f^\star$ is finite on $\mathbb{R}^n$ and is $\frac{1}{\sigma}$-l.c.g.*

Since $g + \mu d$ is $\mu$-sc, Lemma 1 implies $g^\star_\mu$ is $\frac{1}{\mu}$-*l.c.g.* By chain rule, one can show that $g^\star_\mu(A^\top \mathbf{w})$ is $L_\mu$-*l.c.g* where $L_\mu \leq \frac{1}{\mu} \|A\|^2$. Further, the definition of Fenchel dual implies the following uniform deviation bound:

$$g^\star(\mathbf{u}) - \mu D \leq g^\star_\mu(\mathbf{u}) \leq g^\star(\mathbf{u}), \quad \forall\, \mathbf{u} \in \mathbb{R}^n. \tag{16}$$

Hence to find an $\epsilon$ accurate solution to $J(\mathbf{w})$, it suffices to set the maximum deviation $\mu D < \frac{\epsilon}{2}$ (*i.e.* $\mu < \frac{\epsilon}{2D}$), and then find a $\frac{\epsilon}{2}$ accurate solution to $J_\mu$ in (4). Initialize $\mathbf{w}$ to $\mathbf{0}$ and apply Nesterov's accelerated gradient method in (Nesterov, 2007) to $J_\mu$, this takes at most

$$k = \min\left\{\sqrt{\frac{4L_\mu\Delta_0}{\epsilon}}, \ln\frac{L_\mu\Delta_0}{\epsilon} \Big/ \ln\left(1 - \sqrt{\lambda/L_\mu}\right)\right\}$$

number of steps where $\Delta_0 = \frac{1}{2}\|\mathbf{w}^*\|^2$ and $\mathbf{w}^*$ is the minimizer of $J(\mathbf{w})$. Each step involves one gradient query of $g^\star_\mu(A^\top \mathbf{w})$ and some cheap updates. Plugging in $L_\mu \leq \frac{2D}{\epsilon}\|A\|^2$ and using $\ln(1+\delta) \approx \delta$ when $\delta \approx 0$, we get the iteration bound in (5).